\documentclass[letterpaper, 10 pt, conference]{ieeeconf}
\IEEEoverridecommandlockouts
\usepackage{xcolor}   
\usepackage{float}    
\usepackage{caption}  
\usepackage{times}
\usepackage{cite}
\usepackage{graphicx}
\usepackage{url}
\usepackage{booktabs}
\usepackage{xspace}
\newcommand{\sys}{\textbf{LAD-Bench}\xspace}

\title{LADBench: A Benchmark for Logical Fault Detection in Images}

\author{Sahasra Kondapalli$^{*}$, Lara Radovanovic$^{*}$, Aadi Palnitkar, Mingyang Mao, Xiaomin Lin%
\thanks{* Equal Contributors}
\thanks{This work is supported by NSF Award DGE \#2235102 (CyberCorps Scholarship for Service: Cybersecurity Research and Education for Service in Government - CREST), the NVIDIA Academic Grant Program.}
\thanks{Public Dataset: https://huggingface.co/datasets/SahasraK/LADBench}
\thanks{Embodied Robotics and Automation Lab, University of South Florida, Tampa, FL 33610, USA. 
Email: \texttt{\{xlin2\}@usf.edu} 
}
\thanks{© 2026 IEEE. Personal use of this material is permitted. Permission from IEEE must be obtained for all other uses, including reprinting/republishing this material for advertising or promotional purposes, collecting new collected works for resale or redistribution to servers or lists, or reuse of any copyrighted component of this work in other works.}
}

\begin{document}
\makeatletter

\g@addto@macro\@maketitle{
\begin{figure}[H]
  \setlength{\linewidth}{\textwidth}
  \setlength{\hsize}{\textwidth}
    \centering
    \includegraphics[width=0.9\textwidth]{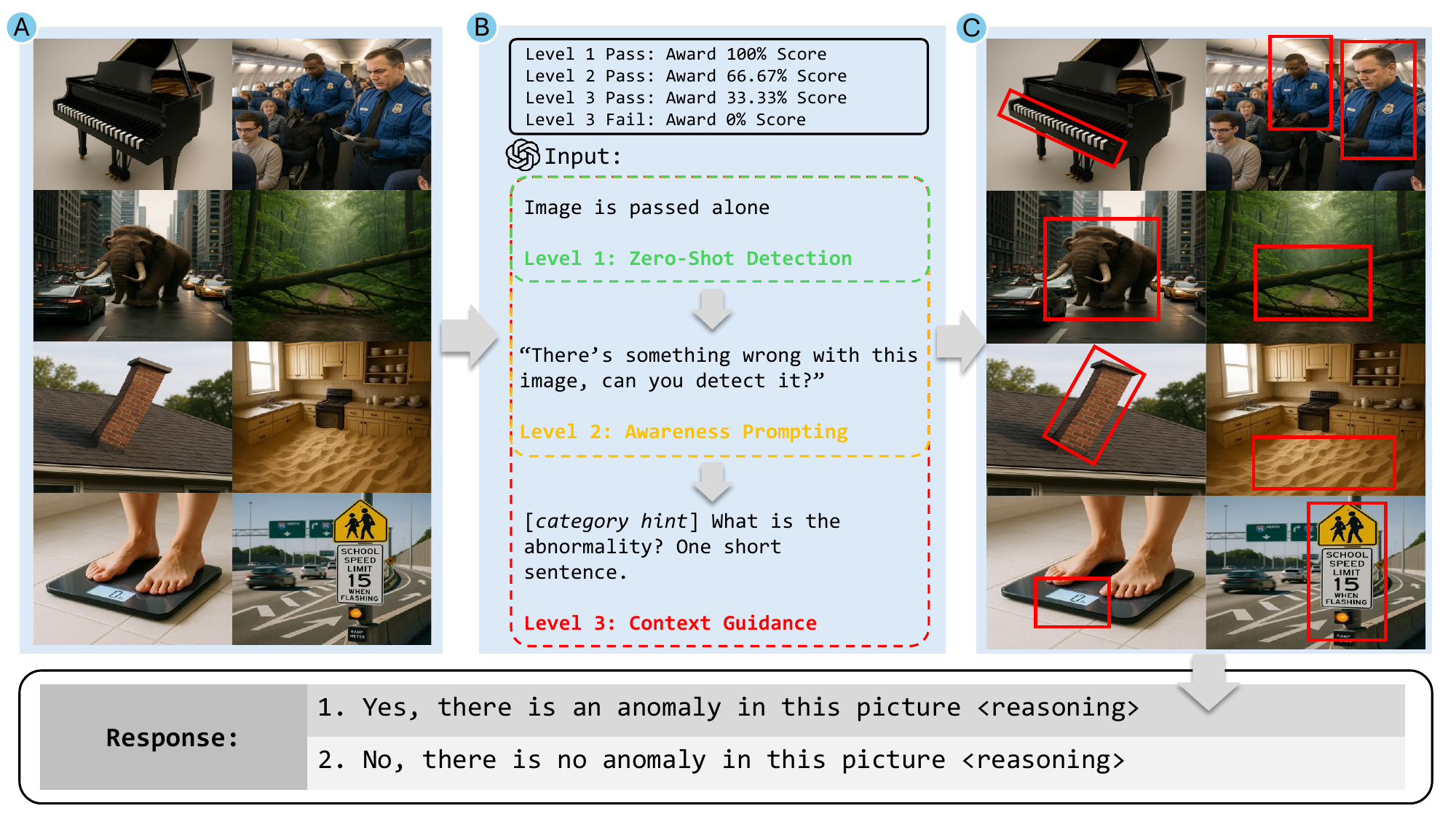}
    \captionsetup{font={footnotesize},labelfont=bf}
    \caption{LAD-Bench evaluation framework and Tiered Prompting Protocol. (A) Left: Synthetic image samples containing various logical anomalies.
(B) Middle: Tiered testing and scoring process. Images initially enter Level 1 (Zero-Shot) for unprompted testing; if the model fails to detect the anomaly, it sequentially advances to the next level to receive additional clues—specifically, Level 2 (explicitly prompting that an anomaly exists in the image) and Level 3 (providing a specific category hint). The score awarded for a successful detection decreases as the level of guidance increases (100\% - 66\% - 33\%).
(C) Right: Ground truth logical errors in the images (highlighted by red boxes), along with the automated evaluation of the model's final output (diagnosis and reasoning process).
}
    \vspace{-10mm}
    \label{fig:Banner}
    \end{figure}
}

\maketitle

\addtocounter{figure}{-1} 
\thispagestyle{empty}
\pagestyle{empty}

\begin{abstract}
Large Vision Language Models (VLMs) excel at visual question answering and semantic grounding, but their capacity for autonomous logical reasoning remains underexplored. Existing anomaly benchmarks emphasize visual errors or direct prompting rather than the physical and social common sense needed for open-world deployment. To address this, we introduce \sys, a benchmark of more than 1,000 curated synthetic images with logical anomalies across four domains: Residential, Urban, Collaborative, and Nature. We further propose a Tiered Prompting Protocol based on progressive disclosure, which measures how much explicit assistance a model needs to localize and reason about a logical fault. Evaluating leading foundation models reveals substantial weaknesses: even the best achieves only 70.11\% overall accuracy, showing that implicit logical fault detection remains unsolved. Crucially, models often fail to identify anomalies even after receiving explicit hints in deeper tiers. By surfacing these limitations in sequential multimodal reasoning, \sys offers a rigorous framework for advancing the safety, reliability, and cognitive alignment of autonomous visual systems. Dataset and Code: https://huggingface.co/datasets/SahasraK/LADBench
\end{abstract}

\section{Introduction}
Vision-Language Models (VLMs) are increasingly employed in various identification tasks in public environments, from threat identification to violation citations. They also find growing use in anomaly detection for industrial and medical applications. These applications involve visual identification, spatial understanding, object isolation and detection, and, arguably most importantly, reasoning. Of all types of reasoning, logical reasoning is of greatest relevance for these applications, as it proves an inherent understanding of the image beyond the data of the pixels. 

Specifically, notable works such as GPT-5V \cite{singh2025openai}, Gemini-Pro-V \cite{team2023gemini}, have demonstrated remarkable reasoning capabilities in image content recognition, and complex scene understanding \cite{liu2024mmbench}. While many claim to have logical understandings, LLMs yet struggle with scenarios wholly outside their training data, hallucinations, and internally strengthened biases. Although strong in surface-level semantics, they remain weak in deeper logical reasoning \cite{zhou2026logic}.

A key limitation in current benchmarks and evaluations of logical reasoning in VLMs is their ability to detect a logical error. Some benchmarks, such as LogicQA \cite{kwon2025logicqa}, try to find logical errors based on predefined constraints, which is useful for Anomaly Detection (AD) applications. Others, such as Salbench \cite{huynh2025vision}, use real images to test the odd-one-out logical capabilities. As such, a number of papers and benchmarks explore the various logical aptitude capabilities of VLMs. An oversight we discovered within these is the applications of a model's logical capabilities in real life scenarios that require a contextual and logical understanding of the scene. 

In this paper, we introduce \sys, a benchmark that contains a dataset of more than 1000 synthetic images, each involving a unique logical anomaly. We define a \textit{logical anomaly} as a scenario that is either physically impossible or significantly less likely than the norm; that is, an aspect of the image that is not normal and constitutes a point of concern to a human observer. These anomalies are curated such that they would be obvious to humans upon seeing the image, but are not direct infractions of any strict logical laws. These include societal abnormalities and a lack of correlation between cause and effect, where one or the other could be missing or mismatched. This reverse reasoning to find the lack of logic, or illogical happenstance, in an image is important for real world, everyday applications, where they are not meticulously trained on predefined objectives.
This capability is foundational to everyday applications, especially in home cameras, consumer grade agents, and embodied robots \cite{grover2026embodied}, and particularly important in autonomous systems, ranging from driving to charting \cite{nekrasov2025spotting}. However, evaluating such reasoning requires analyzing open-ended responses, beyond more objective questions, to gain a more granular understanding of a model's logical and contextual inferential autonomy.

We make \textit{three} primary contributions. First, we introduce a decay-weighted scoring system that rewards autonomous detection, lowering the maximum attainable score as more guidance is given and thus moving beyond binary accuracy to a more representative, weighted measure. Second, we present a comparative analysis across a diverse suite of state-of-the-art open- and closed-source VLMs. Third, we implement a scalable automated grading system using gpt-5-nano \cite{gpt-5-nano}, supported by a human-in-the-loop protocol to certify its accuracy. Together, by quantifying the critical gap in VLM logical reasoning, \sys provides the metrics needed to support reliable autonomous visual systems and drive the next generation of logically consistent multimodal agents.

\section{Related Works}
The effectiveness of neural networks at anomaly detection, in fields of industry and medicine, along with surveillance systems, prompted an increase in research regarding Vision-Language Models in similar contexts. Beyond correlation and analysis, it is to be anticipated that VLMs provide a reasoning aspect to such applications. A number of benchmarks and frameworks were built for VLMs, from applications in video anomaly detection to anomaly classification \cite{xu2025plovad, abdalla2025video}, with some frameworks incorporating logic into anomaly detection in industrial settings \cite{dahmardeh2025nesylad, zhang2025towards}. VLMs also find use in medical image analysis, and anomaly detection therein, for which benchmarks have been defined \cite{shiri2025madclip, huang2024adapting}. Another important, widespread use to VLMs for anomaly detection is in the applications of monitoring or surveillance systems, that need to prioritize safety. A number of benchmarks have also been defined for this particular use case, where the anomalies are usually defined in terms of appearance or temporal context \cite{zhao2025smarthome, delic2025sequential, gani2025vane}. It is evident that a substantial amount of research has been conducted in the applications and capabilities of Vision-Language Models in anomaly detection, particularly in the fields of industry, medicine, and security. 

Beyond anomaly detection, prior work has examined the logical abilities of VLMs in image understanding and reasoning across a range of settings, including comparisons with human performance, aptitude style evaluation, procedural understanding, and domain-specific applied reasoning, and video based temporal reasoning \cite{jin2025logicad, zhang2024gpt, huang2024vbench, mao2025multi, zhou2026logic, bhagwatkar2025cave, kwon2025logicqa, huynh2025vision, mao2026fam}. Most existing benchmarks treat VLMs as human like reasoners and evaluate whether they can interpret image content, infer subtle context, or understand depicted processes. However, they do not directly measure whether a model can identify logical anomalies that violate physical or social common sense without explicit context. \sys addresses this gap by introducing images with clearly illogical aspects that may not break formal rules or obvious visual patterns, yet still appear wrong to humans. This distinction is especially important for real world deployment, where consumer facing embodied systems must reason about everyday situations with broad common sense rather than narrow anomaly cues.

\section{Benchmark}
The \sys benchmark evaluates the capability of Vision-Language Models (VLMs) in detection of logical anomalies, from cause-and-effect mismatch to societal abnormalities, in visual contexts. Our methodology, as illustrated in Fig. ~\ref{fig:Banner}, involves the utilization of guidance level as a metric in identification of logical and common-sense anomalies, allowing for a high-fidelity assessment. This section details the synthetic anomaly dataset, the categories of study, the Tiered Prompting Protocol, and the automated scoring system used for quantification. 
\begin{figure}[t]
    \centering
    \includegraphics[width=\linewidth]{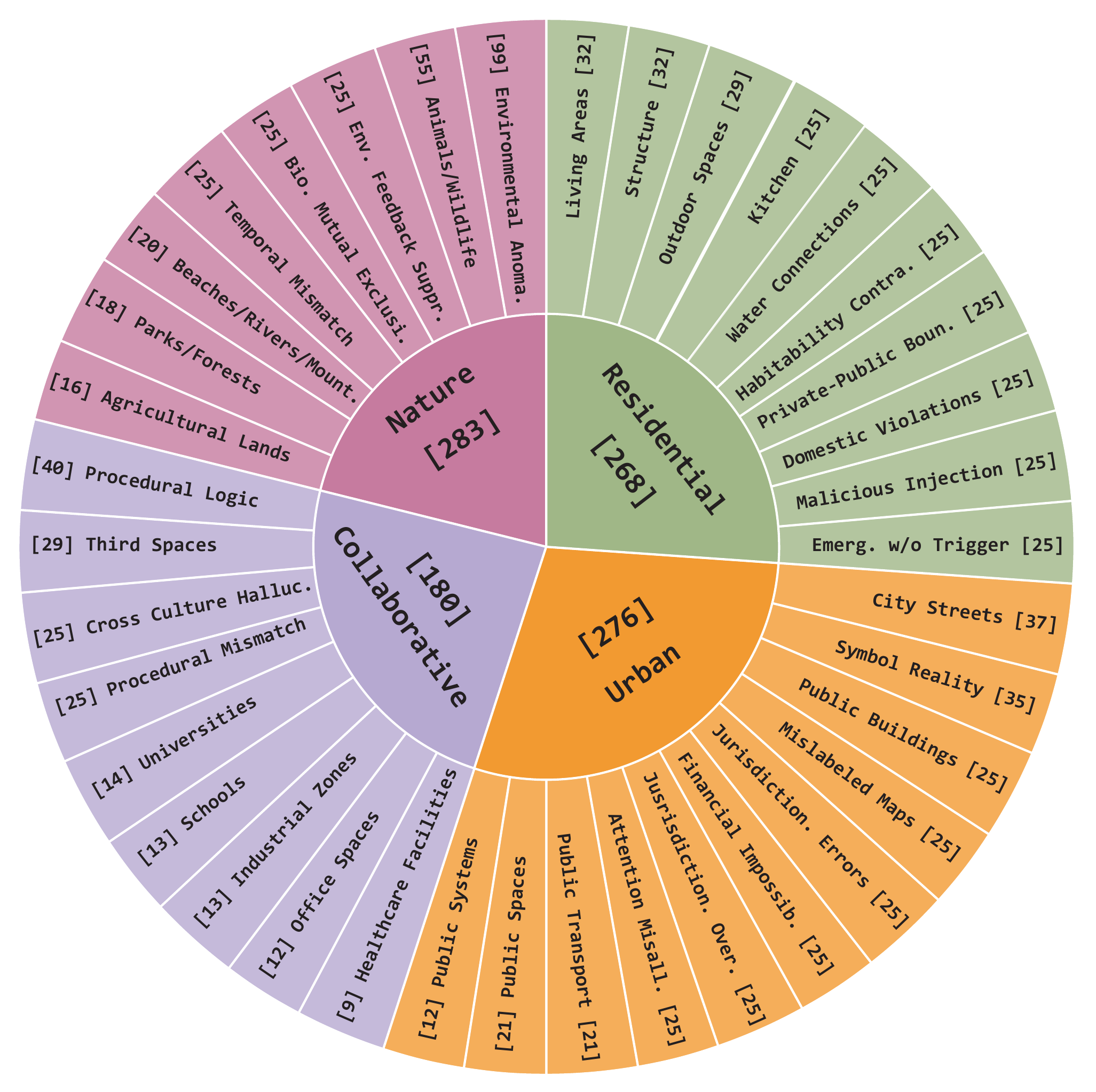}
    \captionsetup{font={footnotesize},labelfont=bf}
    \caption{Distribution of categories and subcategories of anomalous images in the LAD-Bench dataset. The inner ring displays the four core main categories and their total image counts: Nature (283) , Residential (268), Urban (276), and Collaborative (180), representing the primary domains of daily human life. The outer ring lists the specific subcategories and their corresponding image counts.}
    \label{fig:categories}
    \vspace{-7mm}
\end{figure}

\begin{figure*}[t]
    \centering\includegraphics[width=0.95\textwidth]{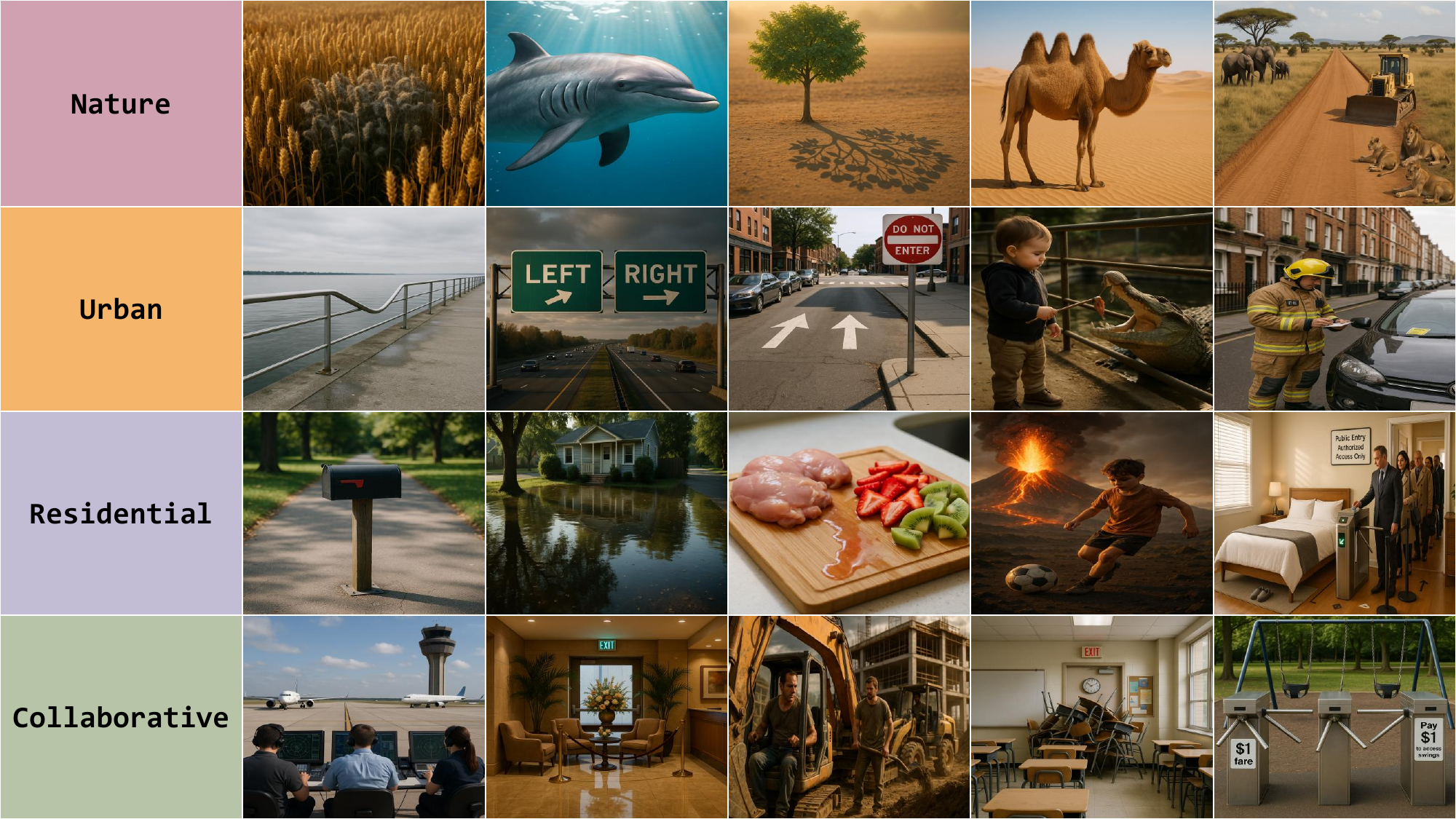}
    \captionsetup{font={footnotesize},labelfont=bf}
    \caption{A collection of example images showing logical anomalies across differentop categories. Nature: Includes scenarios that defy biological or physical laws, such as a dolphin with gills. Urban: Demonstrates public scenarios violating social norms or safety common sense, such as a toddler feeding an alligator at a zoo. Residential: Reflects logical errors in domestic environments, such as a mailbox placed in the middle of a road. Collaborative: Depicts logical breaks in professional processes or group activities, such as ground controllers working directly on the runway.}
    \vspace{-7mm}
    \label{fig:example}
\end{figure*}

\subsection{Dataset}
The dataset for \sys is fully synthetic and contains over 1,000 images generated using OpenAI’s gpt-image-1 model \cite{gpt-image-1}, selected for its image generation quality, speed, and prompt fidelity. Using synthetic images also reduces the risk of overlap with existing training data, since the samples are newly created. This dataset is publicly available at: https://huggingface.co/datasets/SahasraK/LADBench

Each image is organized by both a super category and a sub category. The super category captures the broader daily life domain in which the anomaly appears, while the sub category identifies the specific aspect of that domain being violated. As shown in Fig.~\ref{fig:categories}, the four super categories are Residential, Urban, Collaborative, and Nature, each designed to capture human relevant anomalies in everyday settings.

The logical anomaly is defined by the prompt used to generate the image. Based on human verification of consistency between the prompt and the image, the ground truth label is stored, either as the prompt itself or a slightly modified version of the prompt to better suit the image. This label is used to enable evaluation of the open-ended outputs of the model by comparison with it. 
 
To include a holistic dataset, we independently tested models on 100 normal images taken on the same phone at different times throughout the year, often portraying landscapes and natural scenes. For these, we tested each of them at all three levels and did not consider them for the overall scoring, as the method of testing and the hint provided at the explicit hint level differ considerably, along with the grading method. 

\subsection{Categories}
The categories, as defined with detail in Fig.~\ref{fig:categories}, reflect the overall domains of human life and its settings. 

The \textbf{Residential} category considers domestic settings and activities, including apartments, homes, local streets, residential parks and utilities. An example of this category includes abnormal instances, such as a mailbox set in the middle of the road, as shown in Fig.~\ref{fig:example}. This category is especially important for consumer aimed models, especially ones that may be integrated into robots and machines one keeps at home. Understanding the logic of domestic settings is important to maintain safety and identify threats. 

The \textbf{Urban} category consists of images set in public and urban spaces shared by large numbers of people. It mainly considers organizational faults in these settings, focusing on flaws derived from rules and regulations, both objective and social—such as a zoo letting a toddler feed an alligator, as depicted in Fig.~\ref{fig:example}. This matters for public infrastructure applications, especially the deployment of VLMs in traffic cameras or school bus cameras.

The \textbf{Collaborative} category covers all processes and spaces that involve groups of people. For examples, offices, hospitals, and universities all depend on multiple people and processes. For example, Fig.~\ref{fig:example} depicts ground control of an airport stationed on the runway. Understanding these types of errors makes AI more trustworthy, and ideal for raising alarms before a situation escalates. 

The \textbf{Nature} category considers flaws in naturally occurring things, limited to human-frequented settings such as forest trails, beaches, and other outdoor locations. The flaws range from the logically unreasonable to infrastructure prone to danger. Fig.~\ref{fig:example} illustrates a dolphin with gills, which is biologically improbable. Understanding spatial settings and human safety conventions in outdoor environments is crucial for ensuring trustworthy AI.

\subsection{Tiered Prompting Protocol}
\sys employs the utilization of a Tiered Prompting Protocol to evaluate the models. This consists of three levels, or tiers of guidance. Responses at each level are stored and evaluated by the automated grading module. 

Level 1: Unprompted or Zero-Shot. This level involves sending the model a prompt including only the image, and no companion text prompt. This level evaluates the model's ability to prioritize logical consistency without external stimuli. 

Level 2: Awareness or Prompted. In this level, the model is made aware that there exists a logical inconsistency and anomaly, by the companion text prompt "There is something wrong with this image. Can you detect it?" This level evaluates the model's visual analytical capabilities through enabling awareness.

Level 3: Context Guidance or Hinted. This is the final level, where the model is once again given the image in another request, with a companion prompt that includes a hint to narrow the focus of the model. The prompt is "Hint: {hint}. What is the abnormality? One short sentence." The "hint" here is details on the specific category to which the image belongs. This level evaluates model capabilities at the highest level of guidance, and thus has the least impact on the evaluation. 

\begin{table*}[t]
\centering
\caption{Accuracy comparison across models on the evaluation dataset. Results are reported by category (Residential, Urban, Collaborative, Nature) and reasoning levels (Level 1-3), along with overall weighted and tuned accuracies. Models are compared based on weighted accuracy. Performance generally improves at Level 2 with added hints but declines at Level 3. Toned accuracy reflects adjustments from human-verified false positives and false negatives.}
\label{tab:model-results}
\small
\setlength{\tabcolsep}{5pt}          
\renewcommand{\arraystretch}{1.05}   

\begin{tabular*}{\textwidth}{@{\extracolsep{\fill}} l l c c c c c c}
\toprule
\textbf{Model} & \textbf{Category} &
\multicolumn{3}{c}{\textbf{Levels}} &
\textbf{Weighted Accuracy} & \textbf{Toned Accuracy}\\
\cmidrule(lr){3-5}
& & \textbf{Level One} & \textbf{Level Two} & \textbf{Level Three} & & & \\
\midrule

gpt-5-mini\cite{gpt-5-mini} 
      & Residential    & 41.79\%  & 64.10\%  & 42.86\% & 65.96\% & 60.85\%\\
      & Urban          & 31.52\% & 62.43\%  & 42.25\% \\
      & Collaborative  & 23.33\%  & 62.96\% & 28.00\% \\
      & Nature         & 39.58\%  & 70.76\% & 36.00\% \\
      & Total          & 35.05\% & 64.83\% & 37.39\% \\
      & \textit{No Anomaly}     & 99.00\%  & 15.00\% & 49.00\% \\
\midrule

gpt-5 \cite{gpt-5}
      & Residential    & 51.12\%  & 48.09\%  & 48.53\% & 67.20\% & 70.94\%\\
      & Urban          & 40.94\% & 46.01\%  & 47.73\% \\
      & Collaborative  & 33.89\%  & 46.22\% & 23.44\% \\
      & Nature         & 49.47\%  & 65.03\% & 30.00\% \\
      & Total          & 44.79\% & 51.44\% & 38.89\% \\
      & \textit{No Anomaly}      & 100.00\%  & 10.00\% & 99.00\% \\
\midrule

claude-sonnet-4-6 \cite{claude}
      & Residential    & 45.52\%  & 52.74\%  & 44.93\% & 70.11\% & 72.98\%\\
      & Urban          & 49.28\% & 60.71\%  & 45.45\% \\
      & Collaborative  & 35.56\%  & 62.07\% & 29.55\% \\
      & Nature         & 48.41\%  & 64.38\% & 30.77\% \\
      & Total          & 45.58\% & 59.85\% & 38.64\% \\
      & \textit{No Anomaly}     & 98.98\%  & 29.59\% & 64.29\% \\
\midrule

gemini-3-flash-preview \cite{gemini}
      & Residential    & 52.61\%  & 32.28\%  & 30.23\% & 67.33\% & 76.63\%\\
      & Urban          & 56.88\% & 27.73\%  & 36.05\% \\
      & Collaborative  & 48.89\%  & 34.78\% & 16.67\% \\
      & Nature         & 56.54\%  & 37.40\% & 32.47\% \\
      & Total          & 54.22\% & 32.97\% & 29.77\% \\
      &\textit{No Anomaly}     & 97.00\%  & 20.00\% & 44.00\% \\
\midrule

grok-4.1-fast-reasoning \cite{grok}
      & Residential    & 32.09\%  & 51.10\%  & 28.09\% & 55.28\% & 65.91\%\\
      & Urban          & 34.06\% & 44.51\%  & 34.65\% \\
      & Collaborative  & 24.44\%  & 38.24\% & 19.05\% \\
      & Nature         & 22.61\%  & 57.53\% & 27.96\% \\
      & Total          & 28.60\% & 48.96\% & 27.79\% \\
      & \textit{No Anomaly}     & 100.00\%  & 46.00\% & 84.00\% \\
\midrule

llama3.2-vision:11B \cite{llama3_2_vision}
      & Residential    & 3.36\%  & 18.53\%  & 17.54\% & 18.67\% & 22.37\%\\
      & Urban          & 5.43\% & 16.86\%  & 13.82\% \\
      & Collaborative  & 1.67\%  & 14.12\% & 9.21\% \\
      & Nature         & 6.36\%  & 16.98\% & 10.91\% \\
      & Total          & 4.47\% & 16.84\% & 13.13\% \\
      & \textit{No Anomaly}     & 93.00\%  & 34.00\% & 64.00\% \\
\midrule

qwen3-vl:8B \cite{Qwen3-VL}
      & Residential    & 23.88\%  & 20.59\%  & 31.48\% & 43.66\% & 48.48\%\\
      & Urban          & 28.99\% & 21.43\%  & 33.77\% \\
      & Collaborative  & 21.67\%  & 23.40\% & 16.67\% \\
      & Nature         & 31.45\%  & 28.87\% & 26.09\% \\
      & Total          & 27.01\% & 23.54\% & 27.93\% \\
      & \textit{No Anomaly}     & 100.00\%  & 2.22\% & 49.15\% \\
      
\bottomrule
\end{tabular*}
\vspace{-2mm}
\end{table*}

\subsection{Evaluation methods}
The evaluation of \sys uses an automated grading pipeline together with a decay weighted scoring framework, as shown in Fig.~\ref{fig:Banner}. Under the Tiered Prompting Protocol, each image is presented to a model with increasing levels of guidance. The model response is first evaluated at Level 1, and if unsuccessful, the image proceeds to the next level, up to Level 3. Failure at Level 3 is treated as complete failure. To reflect the amount of assistance required, scores are weighted by level: success at Level 1 earns 100\%, Level 2 earns 66.67\%, and Level 3 earns 33.33\%. This framework rewards models that can identify logical anomalies with less guidance while still capturing partial success under more explicit prompting.

\section{Experiment}
\subsection{Models}
We have conducted testing on seven different models, from both open and closed source providers. The closed source models are: gpt-5-mini \cite{gpt-5-mini}, gpt-5 \cite{gpt-5}, claude-sonnet-4-6\cite{claude}, grok-4.1-fast-reasoning \cite{grok}, and gemini-3-flash-preview \cite{gemini}. The open source models are: llama3.2-vision:11B \cite{llama3_2_vision}, qwen3-vl:8B \cite{Qwen3-VL}.
These models were selected due to their prevalence in the consumer market, and ease of accessibility.

\subsection{Response Validation}
Responses were evaluated using LLM-as-a-judge, where we utilized OpenAI's gpt-5-nano \cite{gpt-5-nano} as our grader. After each response is retrieved, it is sent to the grading model, accompanied by the ground truth label and some additional instructions on how to judge responses. The single word output by the grading model is used as the success index. To offset major errors or oversight in grading, a small portion, about 9.83\%, of responses are human evaluated. The subjectivity of open-ended response evaluation for tasks such as this necessitated keeping the group performing human-verification small. The calculations of false positives and false negatives within this sample are then projected to the whole benchmark by model, which lead to a 4.39\% average adjustment.

\section{Results}
Table \ref{tab:model-results} represents the accuracies recorded, from raw categorical accuracies across levels to overall decay-weighted accuracy for the model. Each level considers a sample space limited to the images tested at that level, excluding previous passes. Results are discussed using weighted accuracy rather than toned since it's the least biased. Across closed source models, claude-sonnet-4-6 performed the best with 70.11\% accuracy, while grok-4.1-fast-reasoning performed the worst with 55.28\%. Interestingly, gpt-5-mini, though meant to use fewer resources than gpt-5 and other mainstream models, performed comparatively to gpt-5, with a difference of 1.23\% in accuracy. 

Although claude-sonnet-4-6 achieved high weighted accuracy, it had lower accuracy at Level 1 than gemini-3-flash preview, relying instead on Level 2 hints for most of its success. This suggests a lack of independent logical situational reasoning. A trend among the models is the increase in success rate at Level Two. Averages in Level 1 detections were shown to be at least 6\% lower than Level 2 detections. The information that there was something wrong in the image substantially improved the models' accuracy and success rates. The exception, gemini-3-flash-preview, showed significantly better success rates at the first level than the higher levels. In addition, it performed the best after the projected evaluation adjustment, with an accuracy of 76.63\%.

In the smaller dataset of false anomalies, where regular pictures were passed to the model under a similar framework, but where all images were tested at all levels, a hint like at Level 2 caused increased hallucinations of anomaly where none were present. Models improved at level 3, where the companion prompt suggested an anomaly and prompted them to "look closely," instead of claiming an anomaly as in Level 2, but improvement varied vastly from model to model, ranging from 99\% accuracy to 44\% accuracy. The performance of the open source models, llama3.2-vision:11B and qwen3-vl:8B, is substantially lower than that of the closed source models. The disparity between open and closed source models highlights the necessity for logical reasoning considerations in designing open source models, and the potential for danger and increased risk in real-life implementations.

\section{Conclusion}
Here, we introduce \sys, a comprehensive benchmark that conducts evaluation of the logical capabilities of VLMs. It considers a variety of settings and tests a broad spectrum of logic. These include identifying visible danger, understanding of physics, understanding of social norms and procedures, and public systems. We also introduce a two step evaluation method to ensure accurate assessment. Furthermore, we comprehensively assess five mainstream commercial AI models and two open source models, using the latest and most prevalent and recent versions from major providers. Our results reveal that current models, are not yet ready for deployment in settings that demand reliable logical reasoning and common sense in priority assignment to potential anomalies. This concern is especially acute for edge applications, where smaller, lighter-weight models are expected to be deployed on household robots, consumer agents, and embedded cameras with the expected reliability of a human operator; these are precisely the class of models \sys is designed to test. This metric can be used in model development to ensure more trustworthy and unbiased AI systems. The evaluation provides invaluable insight that encourages and supports future improvements. 

\bibliography{references1}

@article{singh2025openai,
  title={Openai gpt-5 system card},
  author={Singh, Aaditya and Fry, Adam and Perelman, Adam and Tart, Adam and Ganesh, Adi and El-Kishky, Ahmed and McLaughlin, Aidan and Low, Aiden and Ostrow, AJ and Ananthram, Akhila and others},
  journal={arXiv preprint arXiv:2601.03267},
  year={2025}
}

@article{team2023gemini,
  title={Gemini: a family of highly capable multimodal models},
  author={Team, Gemini and Anil, Rohan and Borgeaud, Sebastian and Alayrac, Jean-Baptiste and Yu, Jiahui and Soricut, Radu and Schalkwyk, Johan and Dai, Andrew M and Hauth, Anja and Millican, Katie and others},
  journal={arXiv preprint arXiv:2312.11805},
  year={2023}
}

@inproceedings{liu2024mmbench,
  title={Mmbench: Is your multi-modal model an all-around player?},
  author={Liu, Yuan and Duan, Haodong and Zhang, Yuanhan and Li, Bo and Zhang, Songyang and Zhao, Wangbo and Yuan, Yike and Wang, Jiaqi and He, Conghui and Liu, Ziwei and others},
  booktitle={European conference on computer vision},
  pages={216--233},
  year={2024},
  organization={Springer}
}

@inproceedings{zhou2026logic,
  title={Logic unseen: Revealing the logical blindspots of vision-language models},
  author={Zhou, Yuchen and Tang, Jiayu and Yang, Shuo and Xiao, Xiaoyan and Dai, Yuqin and Yang, Wenhao and Gou, Chao and Xia, Xiaobo and Chua, Tat-Seng},
  booktitle={Proceedings of the AAAI Conference on Artificial Intelligence},
  volume={40},
  number={34},
  pages={29062--29070},
  year={2026}
}

@inproceedings{kwon2025logicqa,
  title={Logicqa: Logical anomaly detection with vision language model generated questions},
  author={Kwon, Yejin and Moon, Daeun and Oh, Youngje and Yoon, Hyunsoo},
  booktitle={Proceedings of the 63rd Annual Meeting of the Association for Computational Linguistics (Volume 6: Industry Track)},
  pages={411--432},
  year={2025}
}

@inproceedings{huynh2025vision,
  title={Vision-Language Models Can't See the Obvious},
  author={Huynh, Ngoc Dung and Le-Khac, Phuc H and Para, Wamiq Reyaz and Singh, Ankit and Narayan, Sanath},
  booktitle={Proceedings of the IEEE/CVF International Conference on Computer Vision},
  pages={24159--24169},
  year={2025}
}

@misc{gpt-5-mini,
  author       = {OpenAI},
  title        = {{gpt-5-mini}},
  year         = {2025},
  note         = {Vision Capabable Large Language Model},
  url          = {https://developers.openai.com/api/docs/models/gpt-5-mini},
  publisher    = {OpenAI},
  howpublished = {Accessed: 3/3/2026}
}

@misc{gpt-image-1,
  author       = {OpenAI},
  title        = {{gpt-image-1}},
  year         = {2025},
  note         = {Vision Capabable Large Language Model},
  url          = {https://developers.openai.com/api/docs/models/gpt-image-1},
  publisher    = {OpenAI},
  howpublished = {Accessed: 3/3/2026}
}

@misc{gpt-5-nano,
  author       = {OpenAI},
  title        = {{gpt-5-nano}},
  year         = {2025},
  note         = {Vision Capabable Large Language Model},
  url          = {https://developers.openai.com/api/docs/models/gpt-5-nano},
  publisher    = {OpenAI},
  howpublished = {Accessed: 3/3/2026}
}

@misc{gpt-5,
  author       = {OpenAI},
  title        = {{gpt-5}},
  year         = {2025},
  note         = {Vision Capabable Large Language Model},
  url          = {https://developers.openai.com/api/docs/models/gpt-5},
  publisher    = {OpenAI},
  howpublished = {Accessed: 3/3/2026}
}

@misc{claude,
  author       = {Anthropic},
  title        = {{claude-sonnet-4-6}},
  year         = {2026},
  note         = {Vision Capabable Large Language Model},
  url          = {https://www.anthropic.com/claude/sonnet},
  publisher    = {Anthropic},
  howpublished = {Accessed: 3/3/2026}
}

@misc{grok,
  author       = {XAI},
  title        = {{grok-4.1-fast-reasoning}},
  year         = {Year},
  note         = {Vision Capabable Large Language Model},
  url          = {https://docs.x.ai/developers/models/grok-4-1-fast-reasoning},
  publisher    = {2025},
  howpublished = {Accessed: 3/3/2026}
}

@misc{gemini,
  author       = {Google},
  title        = {{gemini-3-flash-preview}},
  year         = {2025},
  note         = {Vision Capabable Large Language Model},
  url          = {https://docs.cloud.google.com/vertex-ai/generative-ai/docs/models/gemini/3-flash},
  publisher    = {Google},
  howpublished = {Accessed: 3/3/2026}
}

@article{llama3_2_vision,
  title={The llama 3 herd of models},
  author={Grattafiori, Aaron and Dubey, Abhimanyu and Jauhri, Abhinav and Pandey, Abhinav and Kadian, Abhishek and Al-Dahle, Ahmad and Letman, Aiesha and Mathur, Akhil and Schelten, Alan and Vaughan, Alex and others},
  journal={arXiv preprint arXiv:2407.21783},
  year={2024}
}

@article{Qwen3-VL,
  author={Bai, Shuai et al.},
  title={Qwen3-VL Technical Report},
  year={2025},
  journal={arXiv:2511.21631},
  url={https://arxiv.org/abs/2511.21631}
}

@inproceedings{nekrasov2025spotting,
  title={Spotting the unexpected (STU): A 3D LiDAR dataset for anomaly segmentation in autonomous driving},
  author={Nekrasov, Alexey and Burdorf, Malcolm and Worrall, Stewart and Leibe, Bastian and Perez, Julie Stephany Berrio},
  booktitle={Proceedings of the Computer Vision and Pattern Recognition Conference},
  pages={11875--11885},
  year={2025}
}

@article{xu2025plovad,
  title={Plovad: Prompting vision-language models for open vocabulary video anomaly detection},
  author={Xu, Chenting and Xu, Ke and Jiang, Xinghao and Sun, Tanfeng},
  journal={IEEE Transactions on Circuits and Systems for Video Technology},
  volume={35},
  number={6},
  pages={5925--5938},
  year={2025},
  publisher={IEEE}
}

@inproceedings{dahmardeh2025nesylad,
  title={NeSyLAD: A Neuro-Symbolic Approach for Unsupervised Logical Anomaly Detection},
  author={Dahmardeh, Malihe and Saadatpour, Mohsen and Manigrasso, Francesco and Morra, Lia and Setti, Francesco},
  booktitle={International Conference on Image Analysis and Processing},
  pages={598--610},
  year={2025},
  organization={Springer}
}

@inproceedings{shiri2025madclip,
  title={MadCLIP: few-shot medical anomaly detection with CLIP},
  author={Shiri, Mahshid and Beyan, Cigdem and Murino, Vittorio},
  booktitle={International Conference on Medical Image Computing and Computer-Assisted Intervention},
  pages={416--426},
  year={2025},
  organization={Springer}
}

@inproceedings{huang2024adapting,
  title={Adapting visual-language models for generalizable anomaly detection in medical images},
  author={Huang, Chaoqin and Jiang, Aofan and Feng, Jinghao and Zhang, Ya and Wang, Xinchao and Wang, Yanfeng},
  booktitle={Proceedings of the IEEE/CVF Conference on Computer Vision and Pattern Recognition},
  pages={11375--11385},
  year={2024}
}

@inproceedings{zhao2025smarthome,
  title={Smarthome-bench: A comprehensive benchmark for video anomaly detection in smart homes using multi-modal large language models},
  author={Zhao, Xinyi and Zhang, Congjing and Guo, Pei and Li, Wei and Chen, Lin and Zhao, Chaoyue and Huang, Shuai},
  booktitle={Proceedings of the Computer Vision and Pattern Recognition Conference},
  pages={3975--3985},
  year={2025}
}

@inproceedings{delic2025sequential,
  title={Sequential keypoint density estimator: an overlooked baseline of skeleton-based video anomaly detection},
  author={Deli{\'c}, Anja and Grcic, Matej and {\v{S}}egvi{\'c}, Sini{\v{s}}a},
  booktitle={Proceedings of the IEEE/CVF International Conference on Computer Vision},
  pages={11579--11589},
  year={2025}
}

@inproceedings{gani2025vane,
  title={Vane-bench: Video anomaly evaluation benchmark for conversational lmms},
  author={Gani, Hanan and Bharadwaj, Rohit and Naseer, Muzammal and Khan, Fahad Shahbaz and Khan, Salman},
  booktitle={Findings of the Association for Computational Linguistics: NAACL 2025},
  pages={3123--3140},
  year={2025}
}

@inproceedings{jin2025logicad,
  title={Logicad: Explainable anomaly detection via vlm-based text feature extraction},
  author={Jin, Er and Feng, Qihui and Mou, Yongli and Lakemeyer, Gerhard and Decker, Stefan and Simons, Oliver and Stegmaier, Johannes},
  booktitle={Proceedings of the AAAI Conference on Artificial Intelligence},
  volume={39},
  number={4},
  pages={4129--4137},
  year={2025}
}

@inproceedings{zhang2024gpt,
  title={Gpt-4v-ad: Exploring grounding potential of vqa-oriented gpt-4v for zero-shot anomaly detection},
  author={Zhang, Jiangning and He, Haoyang and Chen, Xuhai and Xue, Zhucun and Wang, Yabiao and Wang, Chengjie and Xie, Lei and Liu, Yong},
  booktitle={International Joint Conference on Artificial Intelligence},
  pages={3--16},
  year={2024},
  organization={Springer}
}

@inproceedings{huang2024vbench,
  title={Vbench: Comprehensive benchmark suite for video generative models},
  author={Huang, Ziqi and He, Yinan and Yu, Jiashuo and Zhang, Fan and Si, Chenyang and Jiang, Yuming and Zhang, Yuanhan and Wu, Tianxing and Jin, Qingyang and Chanpaisit, Nattapol and others},
  booktitle={Proceedings of the IEEE/CVF Conference on Computer Vision and Pattern Recognition},
  pages={21807--21818},
  year={2024}
}

@inproceedings{bhagwatkar2025cave,
  title={CAVE: Detecting and Explaining Commonsense Anomalies in Visual Environments},
  author={Bhagwatkar, Rishika and Montariol, Syrielle and Romanou, Angelika and Borges, Beatriz and Rish, Irina and Bosselut, Antoine},
  booktitle={Proceedings of the 2025 Conference on Empirical Methods in Natural Language Processing},
  pages={27098--27139},
  year={2025}
}

@inproceedings{zhang2025towards,
  title={Towards training-free anomaly detection with vision and language foundation models},
  author={Zhang, Jinjin and Wang, Guodong and Jin, Yizhou and Huang, Di},
  booktitle={Proceedings of the IEEE/CVF Conference on Computer Vision and Pattern Recognition},
  pages={15204--15213},
  year={2025}
}

@article{abdalla2025video,
  title={Video anomaly detection in 10 years: A survey and outlook},
  author={Abdalla, Moshira and Javed, Sajid and Al Radi, Muaz and Ulhaq, Anwaar and Werghi, Naoufel},
  journal={Neural Computing and Applications},
  volume={37},
  number={32},
  pages={26321--26364},
  year={2025},
  publisher={Springer}
}

@inproceedings{mao2025multi,
  title={Multi-RAG: A multimodal retrieval-augmented generation system for adaptive video understanding},
  author={Mao, Mingyang and Perez-Cabarcas, Mariela M and Kallakuri, Utteja and Waytowich, Nicholas R and Lin, Xiaomin and Mohsenin, Tinoosh},
  booktitle={Pacific-Asia Conference on Knowledge Discovery and Data Mining},
  pages={},
  year={2026},
  organization={Springer}
}

@inproceedings{grover2026embodied,
  title={Embodied foundation models at the edge: A survey of deployment constraints and mitigation strategies},
  author={Grover, Utkarsh and Ranjan, Ravi and Mao, Mingyang and Dong, Trung Tien and Praveen, Satvik and Wu, Zhenqi and Chang, J Morris and Mohsenin, Tinoosh and Sheng, Yi and Polyzou, Agoritsa and Lin, Xiaomin},
  booktitle={arXiv preprint arXiv:2603.16952},
  year={2026}
}

@article{mao2026fam,
  title={FAM-Bench: A Multimodal Benchmark for Condition-Aware Food-as-Medicine Reasoning},
  author={Mao, Mingyang and Medisetti, Bhargav Rishi and Grover, Utkarsh and Ibrahim, Tanvir and Li, Wenyan and Zhang, Tingting and Lin, Xiaomin},
  journal={arXiv preprint arXiv:2605.31410},
  year={2026}
}

\end{document}